\documentclass[sigconf]{acmart}

\AtBeginDocument{%
  \providecommand\BibTeX{{%
    \normalfont B\kern-0.5em{\scshape i\kern-0.25em b}\kern-0.8em\TeX}}}

\setcopyright{none}
\renewcommand\footnotetextcopyrightpermission[1]{} 

\usepackage{algorithm}
\usepackage[noend]{algpseudocode}
\usepackage{hyperref}
\usepackage{multirow}
\usepackage{soul}
\usepackage{amsmath}
\usepackage{textcomp}
\usepackage{subcaption}

\definecolor{flatdarkgray}{HTML}{7F8C8D}
\definecolor{flatgray}{HTML}{BDC3C7}
\definecolor{flatred}{HTML}{C0392B}
\definecolor{flatorange}{HTML}{D35400}
\definecolor{flatyellow}{HTML}{F39C12}
\definecolor{flatdenim}{HTML}{2C3E50}
\definecolor{flatpurple}{HTML}{8E44AD}
\definecolor{flatblue}{HTML}{2980B9}
\definecolor{flatgreen}{HTML}{27AE60}
\definecolor{flatcyan}{HTML}{16A085}
\definecolor{flatdarkgrayalt}{HTML}{95A5A6}
\definecolor{flatgrayalt}{HTML}{ECF0F1}
\definecolor{flatredalt}{HTML}{E74C3C}
\definecolor{flatorangealt}{HTML}{E67E22}
\definecolor{flatyellowalt}{HTML}{F1C40F}
\definecolor{flatdenimalt}{HTML}{34495E}
\definecolor{flatpurplealt}{HTML}{9B59B6}
\definecolor{flatbluealt}{HTML}{3498DB}
\definecolor{flatgreenalt}{HTML}{2ECC71}
\definecolor{flatcyanalt}{HTML}{1ABC9C}

\newtoggle{showtodos}
\togglefalse{showtodos}
\newcommand{\todocolor}[3]{\iftoggle{showtodos}{\textcolor{#1}{\textbf{#2: #3}}}{}}

\newcommand{\eat}[1]{}

\newcommand{\jwendt}[1]{\todocolor{flatpurple}{jwendt}{#1}}
\newcommand{\lucyxie}[1]{\todocolor{flatgreen}{lucyxie}{#1}}
\newcommand{\tata}[1]{\todocolor{flatblue}{tata}{#1}}
\newcommand{\yichaojoey}[1]{\todocolor{flatred}{yichaojoey}{#1}}
\newcommand{\sethebner}[1]{\todocolor{flatyellowalt}{sethebner}{#1}}

\newcommand{\deepform}{FCC Forms}
\newcommand{\investment}{Brokerage Statements}
\newcommand{\mortgage}{Loan Payments}
\newcommand{\paystub}{Earnings}
\newcommand{\utility}{Service Bills}

\newcommand{\fara}{FARA}
\newcommand{\cord}{CORD}

\begin{document}



\title{An Augmentation Strategy for Visually Rich Documents}

\author{Jing Xie}
\affiliation{%
  \institution{Google}  
  \city{Mountain View}
  \country{USA}
}
\email{lucyxie@google.com}

\author{James B. Wendt}
\affiliation{%
  \institution{Google}  
  \city{Mountain View}
  \country{USA}
}
\email{jwendt@google.com}

\author{Yichao Zhou}
\affiliation{%
  \institution{Google}  
  \city{Mountain View}
  \country{USA}
}
\email{yichaojoey@google.com}

\author{Seth Ebner}
\authornote{Work done while at Google.}
\affiliation{%
  \institution{Johns Hopkins University}  
  \city{Baltimore}
  \country{USA}
}
\email{seth@cs.jhu.edu}

\author{Sandeep Tata}
\affiliation{%
  \institution{Google}  
  \city{Mountain View}
  \country{USA}
}
\email{tata@google.com}



\begin{abstract}


Many business workflows require extracting important fields from form-like documents (e.g. bank statements, bills of lading, purchase orders, etc.). Recent techniques for automating this task work well only when trained with large datasets. In this work we propose a novel data augmentation technique to improve performance when training data is scarce, e.g. 10--250 documents. Our technique, which we call FieldSwap, works by swapping out the key phrases of a source field with the key phrases of a target field to generate new synthetic examples of the target field for use in training. We demonstrate that this approach can yield 1--7 F1 point improvements in extraction performance.

\end{abstract}




\maketitle
\pagestyle{plain}  

\section{Introduction}

Visually rich documents like invoices, receipts, paystubs, insurance statements, and tax forms are pervasive in business workflows. Processing these documents continues to involve manually extracting relevant structured information, which is both tedious and error-prone. Consequently, several recent papers have tackled the problem of automatically extracting structured information from such documents \citep{lee2022formnet, garncarek2021lambert, xu2020layoutlmv2, Wu2018FonduerKB, Sarkhel2019VisualSF}. Given a target document type with an associated set of fields of interest, as well as a set of human-annotated training documents, these systems learn to automatically extract the values for these fields from unseen documents of the same type.

While recent models have shown impressive performance~\cite{jaume2019,park2019cord,huang2019icdar2019,stanislawek2021kleister}, a major hurdle in the development of high-quality extraction models is the large cost of acquiring and annotating training documents. 
In this paper, we examine the question of improving the data efficiency for this task especially when only a small number of labeled training documents is available. We propose a novel data augmentation technique called {\em FieldSwap} loosely inspired by the success of mixing approaches in the image domain~\cite{humza2021}. We exploit the observation that most target fields in a document are indicated by a short key phrase (like ``Amount Due'' for the total due in an invoice or ``SSN'' for the Social Security number in a US tax form). FieldSwap generates a synthetic example for a target field $T$ using an example for a source field $S$ and replacing the key phrase associated with $S$ with the phrase for $T$.

\begin{figure}[t]
\includegraphics[width=.95\linewidth]{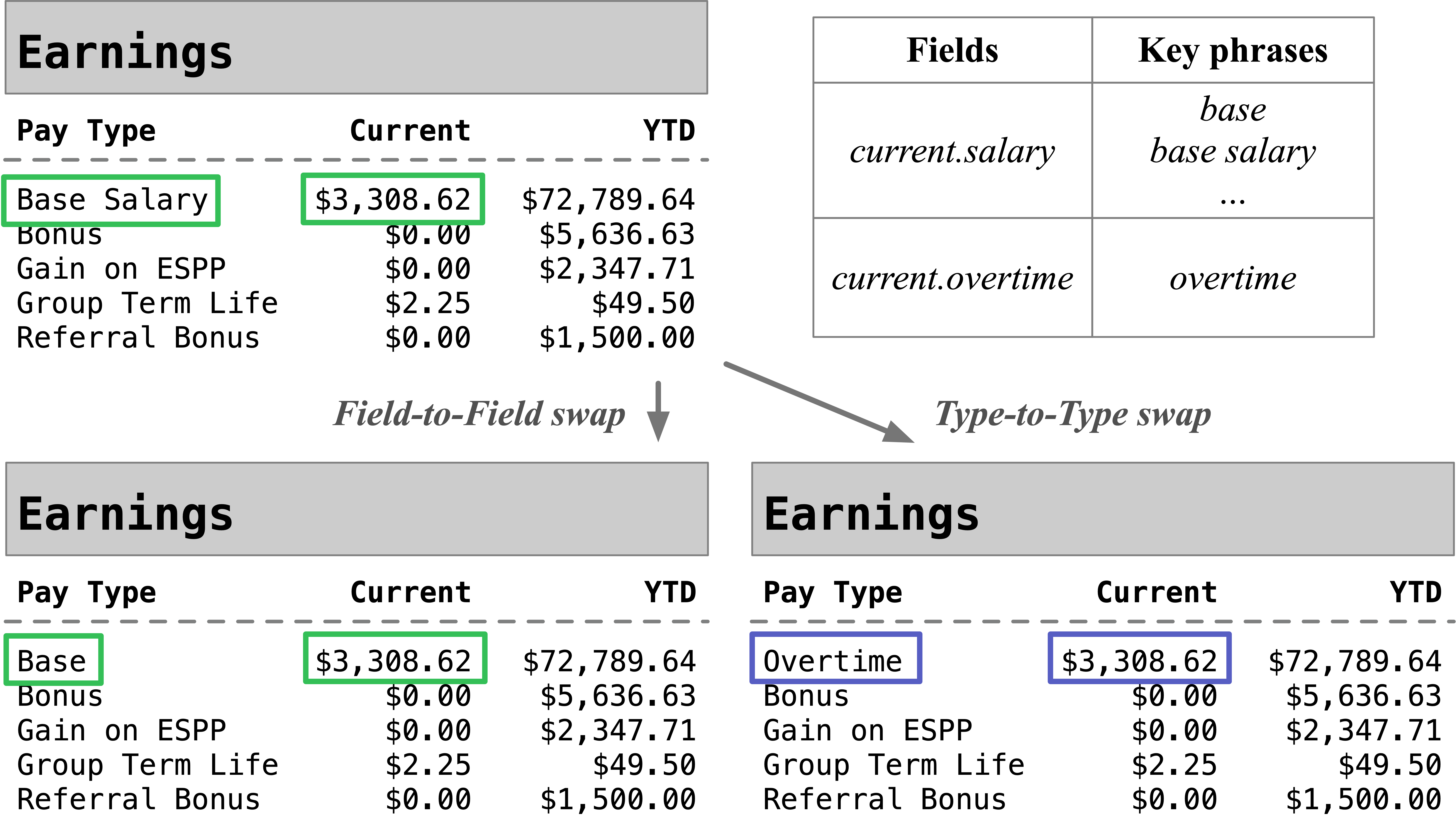}
\centering
\caption{Example of FieldSwap on a paystub document. The source field is \textit{current.salary} with key phrase ``Base Salary''. In the Field-to-Field swap, the phrase is replaced with ``Base'', another key phrase of field \textit{current.salary}, and the label is kept. In the Type-to-Type swap, the phrase is replaced with ``Overtime'', the key phrase of a new target field, \textit{current.overtime}.}
\label{fig:fieldswap_example}
\end{figure}

\tata{\autoref{fig:fieldswap_example} describes Type-to-Type and Field-to-Field examples. We don't discuss that in the introduction (that might actually be too much detail). We should either simplify this figure (preferred) or discuss those ideas here. OK if we want to defer this change to KDD.}
\autoref{fig:fieldswap_example} illustrates FieldSwap on a snippet of a paystub document with typical fields like salary, bonus, and overtime amounts for the pay period and for the year-to-date. As the example in \autoref{fig:fieldswap_example} shows, when the source field $S$ is $current.salary$ (with value \$3,308.62) with the key phrase ``Base Salary'' and the target field $T$ is $current.overtime$ with the key phrase ``Overtime'', the {\em FieldSwap} algorithm generates a synthetic example for the target field. So long as we correctly identify the key phrases associated with the fields $S$ and $T$, generating these synthetic examples serves to regularize the model by presenting the field in different surrounding contexts. Note that the choice of source-target pairs matters. If we instead chose the source field $S$ to be the $ytd.salary$ (with value \$72,789.64) with the same key phrase, the substitution of the key phrase would create an instance of $ytd.overtime$, which is {\em not} a good synthetic example for the field $T = current.overtime$.

The key questions are: (a) how do we infer the key phrases corresponding to each field for a given document type, and (b) what field pairs should be considered for generating these synthetic examples? We show that having a human expert supply a few key phrases works surprisingly well. The challenge then becomes one of inferring these key phrases automatically when only a small number of labeled examples are present. We show that a small model pre-trained for an extraction task on an out-of-domain corpus can be effectively used to identify the key phrases. We also show that simply considering all pairs of fields of the same type can work quite effectively. Experiments on a diverse set of corpora show that depending on the document-type, FieldSwap can produce an improvement of 1--7 F1 points. For context, novel architecture and pre-training objectives in this space resulted in increases of 1--1.5 F1 points~\cite{huang2022layoutlmv3}. We believe this is an exciting step towards better data efficiency in extraction tasks for visually-rich documents that is orthogonal to larger models and larger pre-training corpora.

We make the following contributions in this paper:
\begin{itemize}
    \item We introduce a data augmentation strategy called {\em FieldSwap} that generates new examples for a field $T$ using examples from another field $S$.
    To our knowledge, this is the first instance of a data augmentation strategy designed specifically for visually rich documents.
    \item We present simple algorithms for automatically inferring key phrases and field pairs for generating synthetic examples.
    \item Through experiments on several real-world datasets, we show that {\em FieldSwap} is effective---improving average F1 scores by 1--7 points completely automatically even with small training sets (10--250 documents). 
    \item With simple human inputs like key phrases, we observe improvement up to 14.6 F1 points for some document types. We also present a careful ablation study showing the contribution of each of our design choices.
\end{itemize}
\section{FieldSwap}
\label{sec:field-swap}

FieldSwap exploits the property that form fields are very often indicated by key phrases \cite{majumder-etal-2020-representation}. For example, the \textit{total due} field on an invoice document is often designated by phrases such as ``total'' or ``amount due'', among others. We leverage this observation to generate synthetic examples by taking an instance of a source field, $S$, swapping its key phrase for a key phrase of an intended target field, $T$, and relabeling the instance as an example of the target field. Note that we create a single augmented example for a target field at a time; we do not generate entire synthetic documents with all fields labeled.

This augmentation is governed by two inputs:

\begin{enumerate}
    \item The set of valid key phrases for each field that can be swapped. For example, ``total'' and ``amount due'' are valid key phrases for a \textit{total due} field in invoices.
    \item A list of source-to-target field pairs for which key phrases can be swapped and result in a valid synthetic example for the target field.
\end{enumerate}
These settings can be specified manually by a human or they can be inferred automatically. We find that manually specifying these settings works surprising well (see results in Section \ref{sec:results}). The main challenge is in automatically inferring these settings using only a few examples from a small dataset. Below, we present methods for automatically inferring key phrases and field mappings.

\subsection{Preliminaries}
\label{sec:preliminaries}

FieldSwap, at a high level, is intended to be agnostic to the model architecture for which its augmentations might be used for training. However, there are implementation details that are particular to different model classes.

For model classes that operate at the document-level, such as sequence labeling approaches \cite{xu2020layoutlmv2, lee2022formnet}, there are a number of constraints that complicate the implementation of the FieldSwap augmentation. For example, since field values are also read as inputs to these models, we must also replace both field values along with field key phrases in source examples, lest we introduce inconsistent examples---for example, the values of fields such as \textit{tax due} and \textit{total due} have different relative magnitudes, which should be preserved. Furthermore, since these model classes operate at the document-level, the augmentations must also preserve important document-level semantics. For example, augmented documents must be consistent with the document type's schema---if a field should occur only once in a document, we must ensure FieldSwap does not introduce a second instance. Accommodating these kinds of constraints leads to even more questions---e.g., must augmentations only be made when two fields can act as both sources and targets for one another? Must both fields always be present on the same document in order to perform an augmentation? How might we generate multiple augmentations on a single document?

In this work, we build an implementation of FieldSwap around the binary classifier architecture described in \cite{majumder-etal-2020-representation} as it more easily lends itself to this augmentation strategy than sequence labeling approaches do, since it operates at the candidate-level. In this architecture (as shown in \autoref{fig:glean}), the extraction problem is broken down into two stages:

\begin{enumerate}
    \item Candidate generation: Base types are extracted from an OCR'ed document using common off-the-shelf annotators, such as number annotators, date annotators, and address annotators.
    \item Candidate classification: A binary classifier determines which candidates from the previous stage are the field(s) in question from the document type's schema.
\end{enumerate}

The aforementioned issues that befall sequence labeling architectures are not present in the candidate-level approach since each candidate is treated independently. In fact, the source candidate and target candidate may not even belong to the same document.

A candidate is represented by three features: (i) its position on the page, (ii) its neighboring tokens, and (iii) the relative positions of those neighboring tokens to the candidate. Note that the actual value of the candidate is explicitly ignored. This representation is easy to manipulate for the purposes of FieldSwap since we need only find the key phrase in the neighborhood feature of a source field, replace it with the key phrase of a target field, and relabel the example as the target field.

Thus, we opted to explore FieldSwap first with the candidate-based architecture. We found that this was sufficient to outperform even the state of the art sequence labeling architectures in low data settings. 
For example, \citet{wang2022ABC} evaluated FormNet~\citep{lee2022formnet} with a 10-shot learning task on FCC Forms and achieved $55.15$ Macro-F1 score, which is $11$ points worse than this candidate-based model architecture (as shown in Figure~\ref{fig:field-results}).
We leave it to future work to adapt FieldSwap for the complexities inherent in sequence labeling approaches.

\subsection{Automatically Inferring Key Phrases}
\label{sec:auto-infer-key-phrases}

Recall that, since most fields have short key phrases, only a small number of tokens in a candidate's neighborhood will indicate the candidate's field. We thus define a method for measuring neighbor importance---this metric allows us to detect those tokens that the model pays most attention to when classifying a particular candidate. We aggregate these importance scores across all of the positive examples for a field in the training dataset to infer the set of key phrases for the field. We define a \emph{phrase} as a continuous span of tokens in reading order.

\subsubsection{Neighbor Importance Measurement}

\begin{figure}[t]
\includegraphics[width=\linewidth]{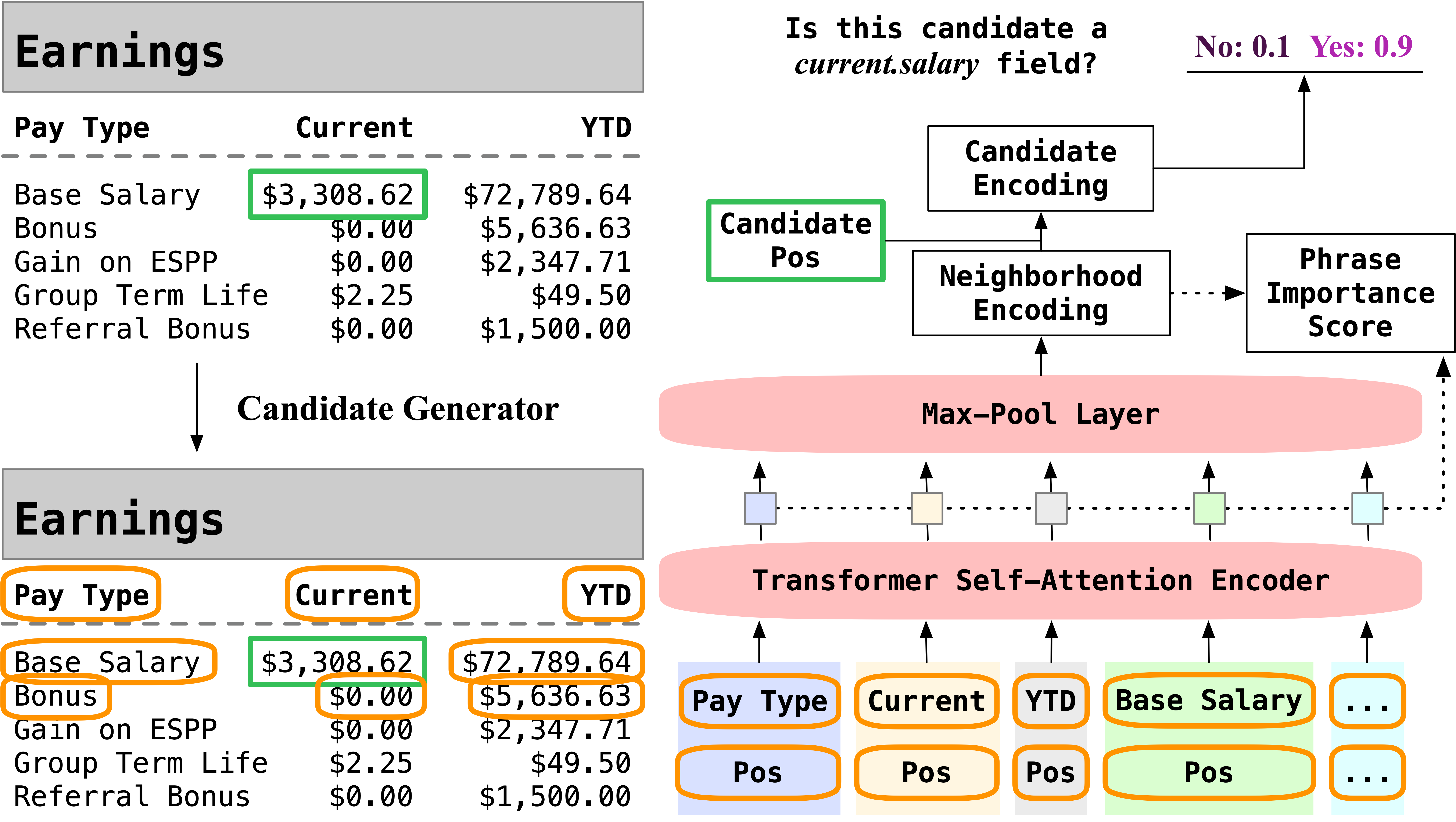}
\centering
\caption{Architecture of the candidate-based extraction model. Neighboring tokens of a \textit{current.salary} candidate (e.g. $\$3,308.62$) are fed into a Transformer-based encoder and a max-pooling layer to generate a neighborhood encoding for measuring the importance of every neighbor phrase. Meanwhile, the neighborhood encoding is concatenated with a candidate position embedding to make a binary prediction for the target field.}
\label{fig:glean}
\end{figure}

We reuse the same model architecture~\citep{majumder-etal-2020-representation} described in Section \ref{sec:preliminaries} for the purposes of measuring neighbor importance. This model learns a field-agnostic representation for each example's neighborhood. It first encodes each neighbor token by concatenating its text embedding and relative position embedding. Then, it employs self-attention and max-pooling to generate the \emph{Neighborhood Encoding} of a candidate. 
To measure how each neighbor contributes to the candidate's representation, we calculate the \emph{cosine similarity} between the model's intermediate output on the encoding of each individual neighbor token and the post max-pooling \emph{Neighborhood Encoding}, as shown in \autoref{fig:glean}. We use a model pretrained on a large out-of-domain dataset for this task. The intuition is that the neighbor's relative position plays a critical role in identifying important neighbors and those positional signals are usually shared across domains. Empirical results show that a pretrained model identifies a reasonable set of important neighbors for each example.

\subsubsection{Inferring Important Phrases per Example}

After obtaining the importance scores of each neighbor for each example, we apply \emph{Sparsemax}~\cite{martins2016softmax} across the importance scores to get a sparse output of the neighbors with non-zero scores. We consider these neighbors as the set of \emph{important neighbor tokens}. We use an OCR service\footnote{\url{https://cloud.google.com/vision}} that detects characters, tokens, and lines in the document. Lines are groups of tokens on the same y-axis that are typically separate from other lines by way of visual features (e.g. vertical bars in a table) or long horizontal stretches of whitespace. Using these signals we form \emph{important phrases} by concatenating all tokens on the same OCR line as long as one token is considered an important neighbor token. \yichaojoey{The \textit{OCR line} may not be an obvious concept for readers so we'd better briefly explain what features that OCR tool provides benefits our FieldSwap algorithm. For simplicity, we may skip the details of this post-processing step.} \jwendt{Official Google OCR documentation on `lines':\\ https://cloud.google.com/document-ai/docs/handle-response, albeit, this is a DocumentAI link, which ties this paper to that product a little too closely, imo. We may want to hold off on this and just describe lines instead.}
This is based on our observation that a key phrase is usually short and lies in a single line. Leveraging OCR lines to infer important phrases also makes the process more tolerant to the model's recall loss on important neighbor tokens. As long as the model finds one important token, we'll be able to infer the longer phrase, if it exists.
We do post-processing of the OCR line inferred phrases by cleaning up any leading and trailing punctuation. We define \emph{phrase importance score} as the average token importance score in the phrase. 

\subsubsection{Aggregation and Ranking by Field}

Once all important phrases and importance scores have been gathered for all positive examples, we aggregate the results by field and phrase. For any field $F$, we use $Score(F, P, C_i)$ to denote the \textit{phrase importance score} for a $F$ candidate $C_i$ that has important phrase $P$ in its neighborhood. We calculate $Importance(F, P) = 1 - \exp( \Sigma_i log(1-Score(F, P, C_i)))$ as the measurement of how $P$ relates to $F$. This measurement prefers phrases with higher importance scores and frequency across $F$ candidates.  We then rank all phrases for each field by their \emph{Importance} and select the top $k$ phrases as the key phrases for the field, where $k$ is a tunable hyperparameter. 

\subsubsection{Fields without Key Phrases}

Not all fields necessarily have key phrases. Fields such as \textit{company name}, \textit{company address}, and \textit{statement date} often appear in the top corners of documents without any specific phrase indicators. When trying to infer key phrases for such fields, the model may output wrong phrases (usually with low importance scores). For example, the model might infer ``LLC'' as a key phrase for the \textit{company address} field since it may often find many company names with ``LLC'' directly above the \textit{company address} field value. However, the values of other fields cannot be part of the key phrase of another field because field values are variable across different documents while key phrases are consistent across the documents (belonging to the same template). To avoid such spurious correlations, we explicitly exclude tokens that are part of the ground truth value for any field in the document. We also set a threshold $\theta$ to filter out any inferred key phrases with importance score below the threshold, where $\theta$ is a tunable hyperparameter.

\subsection{Field Pair Mappings}
\label{sec:field-pair-mappings}
We explored two options to determine which fields can be swapped with each other.

\vspace{5pt}
\noindent\textbf{Field-to-Field swap.} The simplest and most straightforward option is to swap only examples belonging to the same field. In this case, the \emph{source} field, $S$, and \emph{target} field, $T$, denote the same field. With this approach, we are less likely to generate out-of-distribution synthetic examples. The downside is that we are usually unable to generate a sizeable number of synthetic examples unless the field has a lot of key phrase variation. In the extreme case where the field is associated with one key phrase, FieldSwap cannot generate any synthetic examples for this field. In practice, it is the rare fields that we are most interested in augmenting, and these are the fields that benefit the least from this pair mapping approach.

\vspace{5pt}
\noindent\textbf{Type-to-Type swap.} Each field is associated with a general field type such as date, price, number, or address. A simple heuristic is to consider similar fields for FieldSwap, so considering pairs of fields that have the same type is a natural idea to consider. We can generate synthetic examples for a target field (e.g. \textit{salary}) from other same-type fields (e.g. \textit{bonus}, \textit{overtime}) by swapping the key phrases. This approach allows us to generate more synthetic examples for rare fields by utilizing examples from other frequent fields. It also regularizes the model against spurious correlations with nearby non-related text. 

However, we might generate bad synthetic examples if there exist contradictory fields with the same type. For example, \textit{current.bonus} and \textit{ytd.bonus} have the same key phrase ``bonus'', and \textit{current.vacation} and \textit{ytd.vacation} have the same key phrase ``vacation''. FieldSwap would generate contradictory synthetic examples when swapping between the four fields, such as by creating a synthetic \textit{current.vacation} example using a \textit{ytd.bonus} source example. We mitigate the bad effects by ignoring synthetic examples if no neighbor tokens are actually changed and by adding a final training stage in which the model is fine-tuned on the original examples \textit{without augmentations}, which serves to reduce the impact of any out-of-distribution examples we may have introduced by training on bad synthetic examples (details  in Section \ref{sec:training-schedule}).

\vspace{5pt}
\noindent\textbf{All-to-All swap.} We also considered swapping between any pair of fields, but found that this was nearly always worse than type-to-type swaps.

\subsection{Generating Synthetic Examples}
\label{sec:synthetic_examples}
Once the Fieldswap configuration of key phrases and source-to-target pairs is specified (either manually or automatically), we generate synthetic examples. This is done by iterating through all positive examples of the source fields. For each positive example, we iterate through each of the target fields in the source-to-target mapping and attempt to generate one synthetic example for each of the target's key phrases. \sethebner{Include pseudocode if space allows?}

Note that if no phrases in the source example's neighborhood match a source phrase in the configuration, then no synthetic examples are generated from the source example. Furthermore, if, after replacing the source example's key phrase with the target's, we are left with an unchanged neighborhood, we also omit the synthetic example. This helps prevent us from creating semantically incorrect examples, such as the case previously described in Section \ref{sec:field-pair-mappings}, when two fields have the same key phrase but are semantically different (e.g., \textit{current.bonus} vs. \textit{ytd.bonus}). 

When a source phrase and target phrase are exactly the same length in tokens, each source phrase token is replaced with its counterpart target phrase token. When the source phrase is longer than the target phrase, we replace the first $n_t$ tokens, where $n_t$ is the number of tokens in the target phrase, with the target phrase tokens, then PAD the remaining $n_s-n_t$ tokens belonging to the source phrase, where $n_s$ is the number of source phrase tokens. When the source phrase is shorter than the target phrase, we replace the entire set of source phrase tokens with the first $n_s$ target phrase tokens, then replace $n_t-n_s$ of the lowest important tokens in the neighborhood by way of the same importance measurement methods described above. In this scenario we also update the $n_t-n_s$ trailing target phrase tokens' position features to that of the position of the $n_s$th token.

\subsection{Training Schedule}
\label{sec:training-schedule}
We design a training schedule to avoid the model learning poorly from the augmented instances, which may deviate from the original data distribution. This is accomplished by first training the model on all of the augmented instances together with the initial training dataset, then fine-tuning the model on the initial training dataset.  Empirically, we see large improvements made by this training schedule, demonstrating that this fine-tuning step helps the model learn from both augmented and training data while forgetting out-of-distribution information that can hurt performance. We also down-weight the augmented instances in the first training stage so that we can control how aggressively the model learns from the synthetic examples. Detailed experiments are discussed in Section~\ref{sec:ablation}.
\tata{Would be nice to cite a paper for this training schedule idea. Can we repoint to feng-etal-2021-survey?}
\tata{(For the KDD version), adding a figure that summarizes Section 2 (Infer key-phrases, generate synthetic examples, train on T U S, then train on T. Would help. I think some of what Figure 1 does could be pulled out (like the inferred key phrases).}
\section{Experiments}

\subsection{Dataset}
\label{sec:dataset}

\begin{table}[t]
    \centering
    \resizebox{.95\linewidth}{!}{
    \begin{tabular}{c|c|c|c}
    \hline
    Document Type & \# Fields & Train Docs Pool Size & Test Docs \\
        \hline
        \hline
        {\investment} & 18 & 294 & 186 \\
        {\mortgage} & 24 & 2000 & 815 \\
        {\paystub} & 26 & 2000 & 1847 \\
        {\utility} & 25 & 1086 & 183 \\
        {\cord} \cite{park2019cord} & 23  & 800 & 100 \\
        {\deepform} \cite{wang2022ABC} & 13 & 200 & 300 \\
        {\fara} \cite{wang2022ABC} & 6 & 200 & 300 \\

        \hline
    \end{tabular}
    }
    \caption{Datasets. A subset of documents are selected at random from the larger pool to create the training sets for our experiments.}
    \label{tab:dataset}
\end{table}

\begin{table}[t]
    \centering
    \resizebox{.95\linewidth}{!}{
    \begin{tabular}{l|c|c}
    \hline
    Hyperparameter & Range explored & Best performer \\
        \hline
        \hline
        learning rate & 0.0001-0.1 & 0.001 \\
        dropout rate & 0.1-0.5 & 0.1 \\
        batch size & \{64, 128, 256\} & 128 \\
        top-$k$ phrases per field & 1-3 & 3 \\
        importance score threshold $\theta$ & 0-0.3 & 0.2 \\
        downweight ratio &  0.1-1  & 0.4 \\
        
        \hline
    \end{tabular}
    }
    \caption{Hyperparameters.}
    \label{tab:hyperparam}
\end{table}
We evaluate FieldSwap on seven datasets of form-like documents, including three public datasets: \cord\cite{park2019cord}, \deepform\cite{wang2022ABC}, \fara\cite{wang2022ABC}, and four proprietary datasets: \paystub, \investment, \mortgage{} and \utility. Each dataset corresponds to a different document type. Dataset statistics are summarized in \autoref{tab:dataset}. We herein also refer to document types as domains. For each domain, we evaluate on a fixed hold-out test set. We vary the training dataset sizes to plot learning curves.\lucyxie{Bit of a pain, but for the KDD submission we should probably get rid of the 250 doc size and just re-run our experiments at 200 to make this cleaner.}

\subsection{Experimental Setup}
\label{sec:experiment_setup}

In all experiments, we started with a pretrained model that was trained on an out-of-domain document type (invoices) with approximately $5000$ training documents. We fine-tuned on the training examples in the target domain, as described in \cite{gunel2022data}. We tune the hyperparameters using grid search and use the most performant values, listed in \autoref{tab:hyperparam}. When training the model on the target document type, we split the dataset into 80\%-20\% training-validation sets.

We train the model on the target domain in two stages. In the first stage we generate FieldSwap augmentations and append them to the target domain training data, and train for 10 epochs. In the second stage we fine-tune the model on the target domain training data without any synthetic examples for 25 epochs. We stop training early in each stage if the AUC-ROC\sethebner{Does this metric need to be explained?} on the validation set has not improved over the last 3 epochs. We find it typically takes fewer than 10 epochs to trigger early stopping.

We pick the checkpoint from the fine-tuned stage with the best AUC-ROC on the validation split. In order to fully utilize all examples from the sample training set, we use a simple multi-model ensemble technique~\cite{dong2020survey}. We train 3 separate models with the same architecture and the same training schedule but with different seeded training-validation splits and different parameter initialization. We ensemble the 3 models by averaging their outputs. We find simultaneously training and ensembling multiple models can improve the performance substantially.

We evaluate the ensembled model by measuring the end-to-end field extraction performance on the test set using the maximum F1 in the precision-recall curve.\sethebner{Maximum over what? How is the precision-recall curve explored? Is there a hyperparameter being set?} We report the macro average of the F1 scores across all fields.

\subsection{Baseline and Variability}
\label{sec:variability}
We build our baseline model using the same architecture, transfer learning, and multi-model ensemble technique mentioned above. The only difference is that we don't incorporate FieldSwap augmentation in the baseline setting.

In order to capture the inherent variability that may arise in experiments with such small dataset sizes, we repeat our experiments across three different axes. For a given domain and dataset size $N$, we repeat the experiment using (i) 3 different collections of $N$ documents from the domain's large pool of documents (see \autoref{tab:dataset}), (ii) 3 training-validation splits of those $N$ documents, and (iii) 3 differently seeded random initializations of the model's parameters. This amounts to a total of 27 experiments for the given domain and dataset size. Each data point we report on the learning curve corresponds to the median performance across these 27 experiments on the fixed hold-out test set.


\subsection{Human Expert}
\label{sec:oracle}

A human familiar with a given document type can easily provide additional inputs in lieu of additional labeled examples. For instance, a human can provide typical key phrases that indicate a field as well as mark potential pairs of fields that should be used for FieldSwap. This idea draws on the literature in rule-based augmentations where rules are provided in addition to training examples.
We design a \textit{human expert} approach by devising a FieldSwap configuration with human inputs. Instead of relying only on the automatically detected key phrases and field pairs to conduct FieldSwap augmentations, we apply human inputs to protect FieldSwap from some error-prone situations. For key phrases, fields such as \textit{debtor\_name} and \textit{debtor\_address} from the \mortgage{} corpora do not have clear key phrases, we remove these fields entirely in this \textit{human expert} setup. In some cases, particularly rare fields such as \textit{pto\_pay} and \textit{incentive\_pay} from the \paystub{} corpora might not even have positive examples (which is common in the few-shot setting), so we rely on domain knowledge to supply the key phrases. For field pairs, we start with type-to-type field pairs, then prune those that are mostly likely to appear in different tables or sections in the document. 



\subsection{Results}
\label{sec:results}
The experiments are aimed at answering the following four questions: (1) Is FieldSwap effective in its fully automatic version? (2) Does it work better  with human-supplied inputs? (3) How do improvements vary across document types and field types? (4) Which design choices were critical?

\begin{figure*}[t]
\begin{subfigure}{\textwidth}
\includegraphics[width=.95\linewidth]{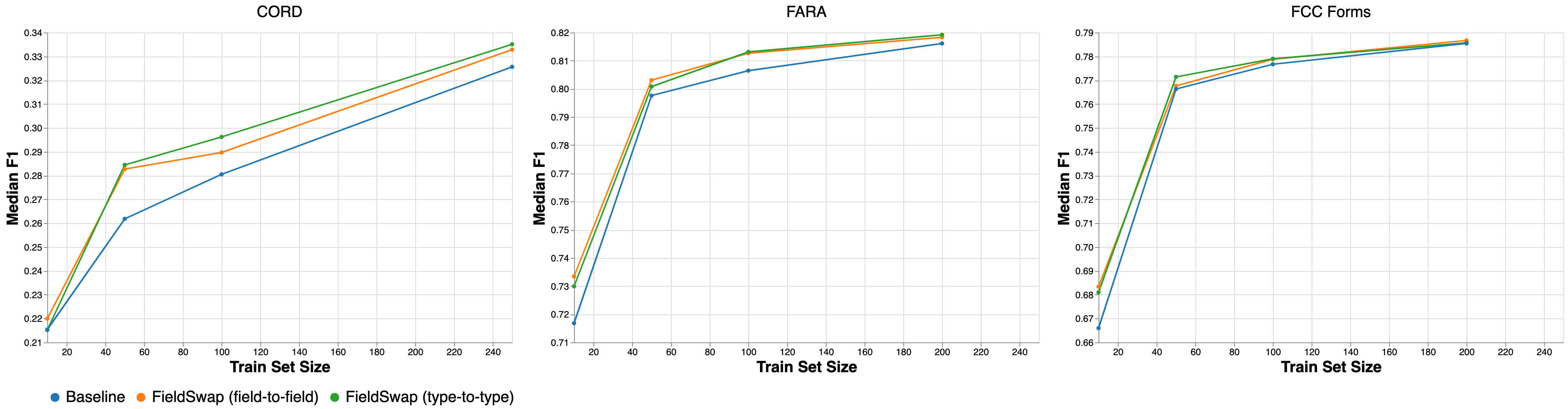}
\centering
\caption{Public Datasets}
\end{subfigure}
\begin{subfigure}{\textwidth}
\includegraphics[width=.95\linewidth]{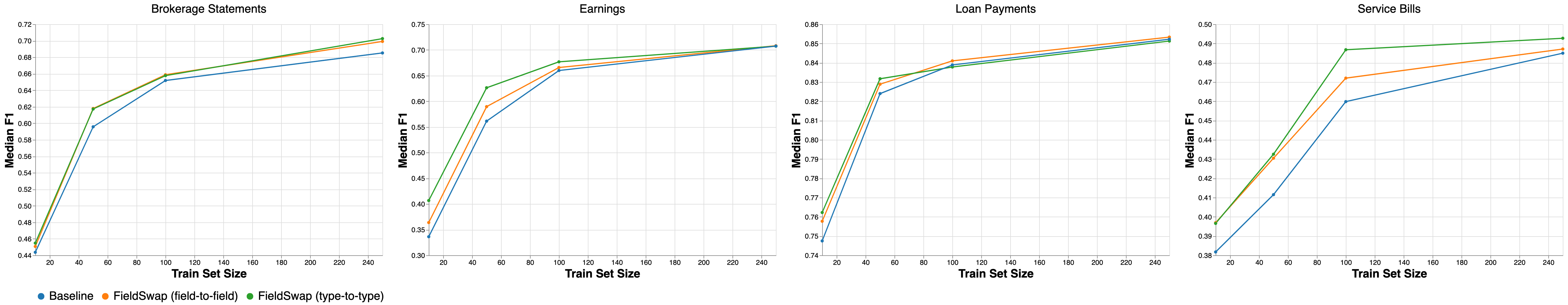}
\centering
\caption{Proprietary Datasets}
\end{subfigure}
\caption{Experiment results on different domains with different train doc sizes. Under each setting, we repeat the experiments with different random seeds as mentioned in Section \ref{sec:variability}. We report the median Macro-F1 scores.\yichaojoey{Figures 3 and 4 can be improved to have the same font size. It's not obvious from the curves how many extra F1 points or percentages were brought by Fieldswap algorithms. Shapes of each curve shall be different (e.g. triangles, squares) so that readers can differentiate between the curves in black-and-white prints too.}}
\label{fig:results}
\end{figure*}

\noindent\textbf{Automatic FieldSwap.} FieldSwap improves macro average F1 by 1-7 points across 6 of the 7 datasets we evaluated on, while the performance for some individual fields improves by as much as 22 F1 points.
The magnitude of improvement varies across document types and even across dataset sizes, but FieldSwap never hurts performance. Most consistently, improvements are most apparent in the lower data regimes ($<100$ documents), and for some document types improvements continue up to larger dataset sizes.

We test both field-to-field and type-to-type field mappings with automatically inferred key phrases. As shown in \autoref{fig:results}, FieldSwap improves extraction performance at small dataset sizes (10--250 documents). Type-to-type FieldSwap typically performs best.

The biggest improvement we observe is on the {\paystub} domain. Compared to other document types, most of the fields in {\paystub} are in tabular format, with similar base type (i.e. numerical values), and have clear and succinct phrase indicators. We believe FieldSwap is most helpful when dealing with document types with such characteristics. Furthermore, since ordering of fields in these tables is unimportant, FieldSwap is particularly well-aligned for augmenting these structures.
That being said, the \paystub{} domain also poses a challenge since many field pairs can easily yield bad synthetic examples, as discussed in Section \ref{sec:field-pair-mappings} (e.g., \textit{current.X} vs. \textit{ytd.X}). Yet, even in the presence of these potentially contradicting pairs the type-to-type mapping still yields the best performance. This demonstrates that the proposed method tolerates a small number of these contradictory synthetic examples. \lucyxie{What intuition can we provide on why this works out?}

FieldSwap is also not useful for all domains.\sethebner{What is this supposed to mean? That not every domain benefits from it?} For example, the {\deepform} corpus contains only a handful of swappable field pairs, namely, name and address types, many of which in fact do not have indicative key phrases. When fields are not marked explicitly with key phrases, FieldSwap should explicitly not generate augmentations. For the most part, our implementation does exactly this, however at times it does fail and find spurious phrases for these fields. However, in this scenarios, the additional safeguards put in place (e.g., omitting ground truth text as described in Section~\ref{sec:synthetic_examples}) helps maintain neutral results throughout the learning curves for these fields.
\sethebner{The plots are useful, but we should also include the exact numbers in a table, especially for future work to compare against. tata: +1, especially an Appendix for the arxiv version makes perfect sense.}

\begin{figure*}[t]
\includegraphics[width=.95\linewidth]{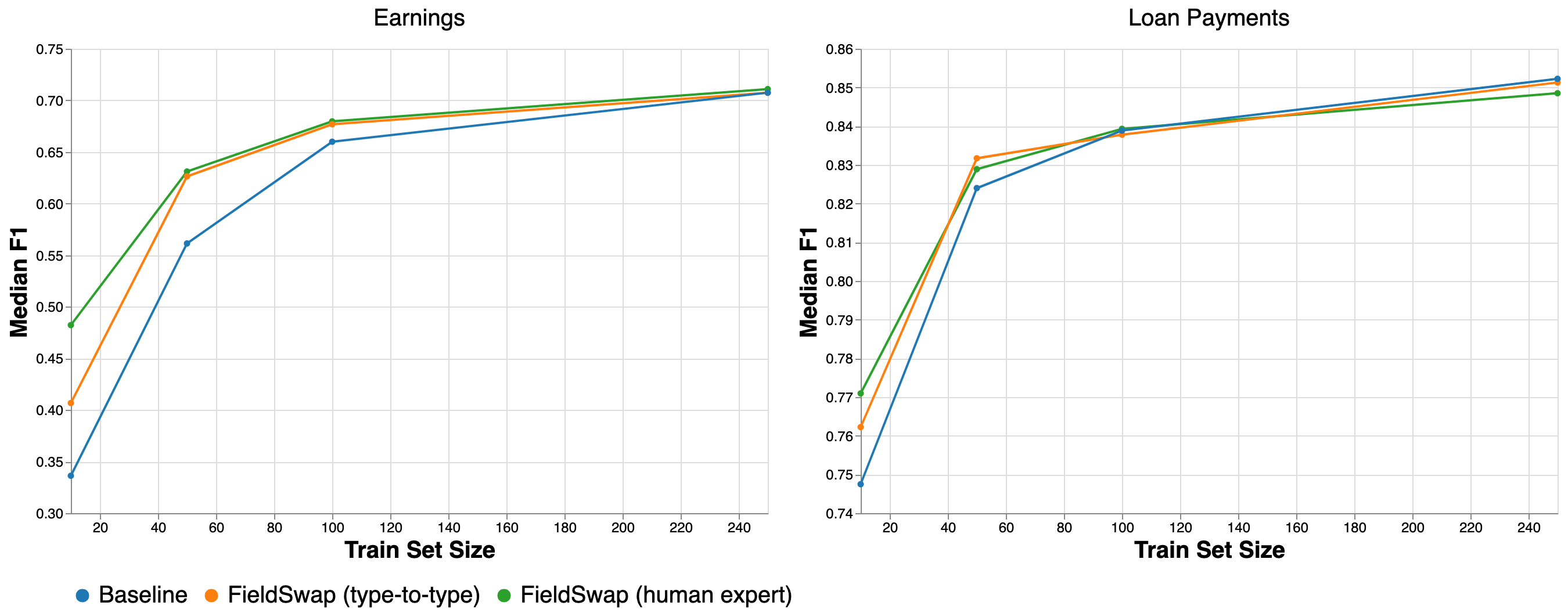}
\centering
\caption{Learning curves of Baseline, FieldSwap (type-to-type) and FieldSwap (human expert) on {\paystub} and {\mortgage} domain.}
\label{fig:oracle}
\end{figure*}

\begin{table}[t]
    \centering
    \resizebox{.95\linewidth}{!}{
    \begin{tabular}{llcccc}
    \toprule
        Domain &                               Field & Frequency &  \shortstack{F1 (FieldSwap,\\ automatic)}  &  \shortstack{F1 (FieldSwap,\\human expert)} &  $\Delta$F1  \\
    \midrule
    \midrule
       \paystub &             ytd.sales\_pay &            3.9\% &                      0.62 &               51.28 &  50.66 \\
        &                ytd.pto\_pay &           15.9\% &                      7.05 &               50.88 &  43.84 \\
        &                 current.sales\_pay &           2.85\% &                      3.28 &               33.71 &  30.43 \\
        &                    current.pto\_pay &            9.5\% &                      7.41 &               35.57 &  28.16 \\
        &                       current.incentive\_pay &            8.3\% &                     15.15 &               31.97 &  16.82 \\
    \hline
     \mortgage &  overdue\_charges &           32.1\% &                     78.66 &               86.86 &   8.20 \\
      &    date\_of\_maturity &          25.55\% &                     46.05 &               50.77 &   4.71 \\
      &         finance\_charges &          32.65\% &                     88.68 &               90.73 &   2.05 \\
    \bottomrule
    \end{tabular}

    }
    \vspace{5pt}
    \caption{Fields with the largest Median F1 score gaps between Automatic FieldSwap and Human Expert FieldSwap when trained on 10 training documents. Frequency refers to the fraction of documents that contain said field in the larger pool of training documents.}
    \label{tab:oracle-gap}
\end{table}

\vspace{5pt}
\noindent\textbf{FieldSwap with Human Expert.}  We ran experiments to answer two questions: (1) Can inputs provided by a human expert be easily incorporated into FieldSwap? (2) How does this compare with the algorithm to automatically infer key phrases?
We compare the performance between automatic FieldSwap with type-to-type mappings and FieldSwap using human expert-curated phrases and mappings. We evaluate these two settings on the {\paystub} and {\mortgage} domains.

In the human expert setting, we obtain the key phrases by inspecting examples in the training set---one of the authors of this paper examined a small number of training documents in each domain and recorded the key phrases they observed for each field. The same person also constructed the field mappings using the method described in Section \ref{sec:oracle}, which avoids contradictory field pairs.

\autoref{fig:oracle} shows that performance can be further improved with inputs from human experts in the low data regime (10 documents). For instance, at the 10 document setting, human inputs result in an F1 score that is nearly 7 points higher on the {\paystub} corpus and about 1 F1 point higher on the {\mortgage} domain. At larger dataset sizes (250 documents), this advantage dissipates and the results are neutral when comparing with the automatic approach.
In \autoref{tab:oracle-gap}, we observe that the gap in the 10 document setting is mostly attributed to rare occurring fields. If the human supplied key phrases that were not present in the small training set, this resulted in a substantial advantage over the automatic approach which has no hope of discovering this key phrase. With larger dataset sizes, the key-phrases provided by the human were discoverable from the data. These results lead us to conclude that inputs from a human expert may be useful for zero-shot or few-shot scenarios.

\begin{figure}[t]
\includegraphics[width=.95\linewidth]{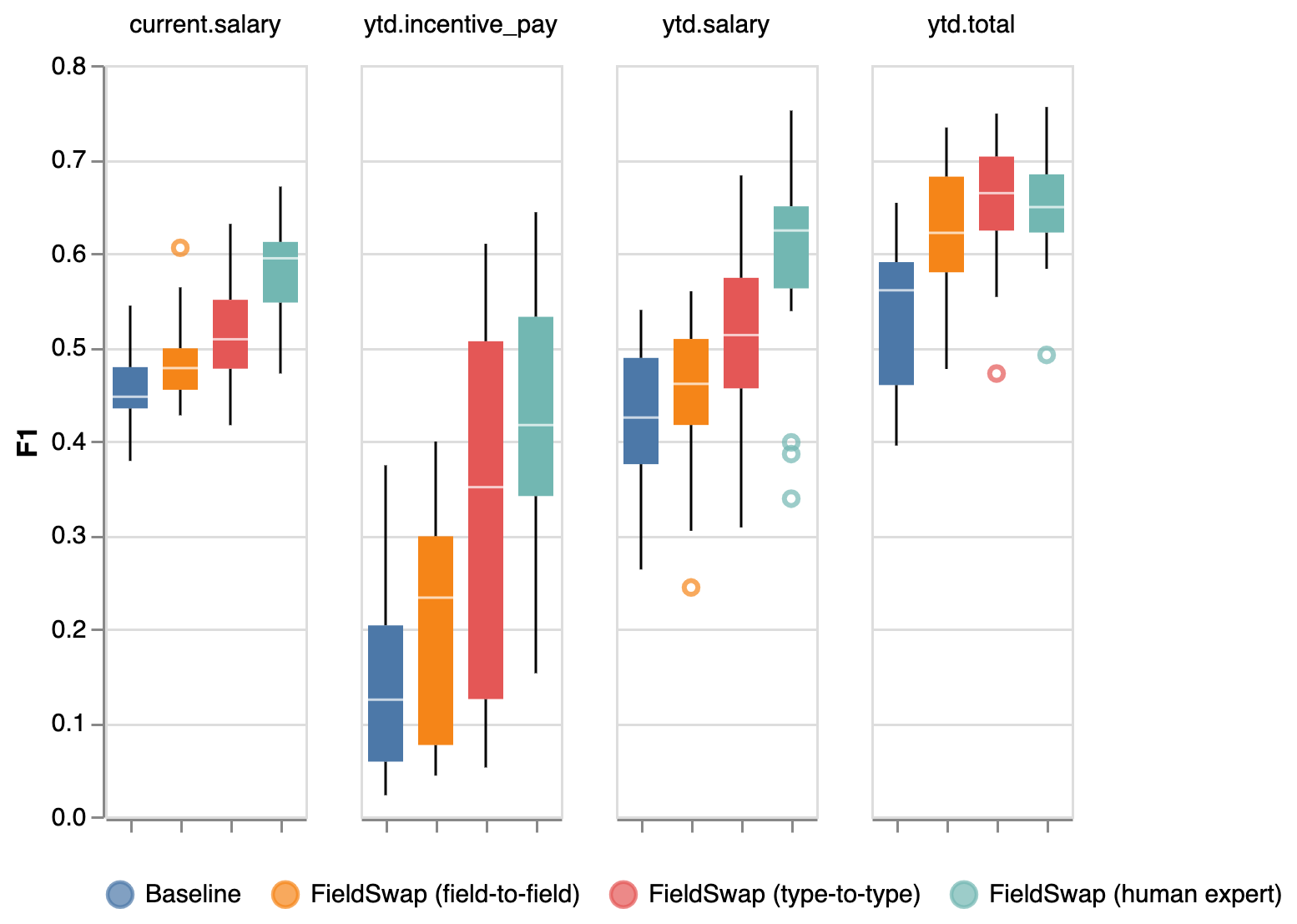}
\centering
\caption{Per-field median F1 at 10-document train set size with standard deviation error bars. \yichaojoey{What the length of each box plot, dots, lines mean in Figure 5 shall be explained in the caption. And readers need to spend a few minutes calculating before they find out why the improvement is 22 F1 points.}}
\label{fig:field-results}
\end{figure}

\vspace{5pt}
\noindent\textbf{Per-field Extraction Performance.}
\autoref{fig:field-results} shows a comparison between the baseline and FieldSwap with different settings on 4 fields in the {\paystub} domain. We find the value of FieldSwap can be particularly impressive in the few-shot setting for some fields, with improvements of 22 F1 points, even without human inputs. In general, we tend to find the largest improvements for the rarer fields.
Note that at small dataset sizes, the particular examples that are present in the training data can have a large impact on downstream evaluation metrics. For example, consider a rare field that occurs in $\sim$20\% of documents, and a dataset size of 10 documents---a difference of 1 or 2 occurrences can mean a 50\% difference in available training signal.

\begin{figure*}[t]
\includegraphics[width=.95\linewidth]{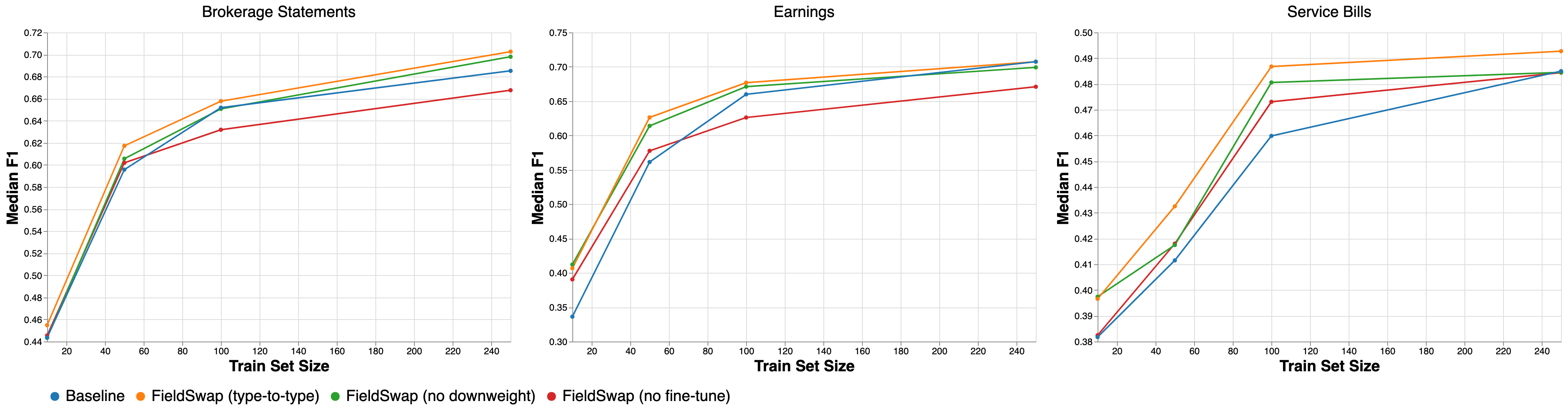}
\centering
\caption{Ablation study. FieldSwap (type-to-type) has both downweighting and fine-tune features turned on. FieldSwap (no downweight) only enables fine-tune stage on the initial dataset without downweighting augmented instances. FieldSwap (no fine-tune) only downweights augmented instances without the additional fine-tuning stage.}
\label{fig:ablation}
\end{figure*}

\subsection{Ablations}
\label{sec:ablation}
\autoref{fig:ablation} presents an ablation study to understand the impact of the more important design choices we made, in particular, the training schedule and downweighting of synthetic examples, both described in Section \ref{sec:training-schedule}.

We use auto-inferred key phrases and type-to-type swap in all FieldSwap settings and also compare the results against the baseline. For FieldSwap, we disable one feature at a time to test the effectiveness of down-weighting the augmented instances and fine-tuning on the initial training dataset. Results show that both features are helpful. We also notice that the performance of FieldSwap could drop below the baseline as the training set size increases when we disable fine-tuning. The largest drop is on the {\paystub} domain when the training set reaches 50 documents. Recall that the {\paystub} domain contains many contradictory fields. As the training set size increases, FieldSwap with type-to-type swap has more chances to generate contradictory synthetic examples that could hurt the model. We find that the additional fine-tuning stage on the initial training dataset is effective at mitigating such bad effects. 

\section{Related Work}

\subsection{Data Augmentation}
Data augmentation is a class of techniques for acquiring additional data automatically. Two main categories of data augmentation are rule-based and model-based techniques, which use hard-coded data transformations or pre-trained models (typically language models), respectively. Rule-based techniques---such as EDA \cite{wei-zou-2019-eda}---are easier to implement but have limited benefit, whereas model-based techniques---such as back-translation \cite{sennrich-etal-2016-improving} and example extrapolation \cite{lee2021neural}---are more difficult to develop but offer greater benefit \cite{feng-etal-2021-survey}. FieldSwap contains elements of both categories, as it changes (possibly automatically inferred) key phrases based on a set of swap rules.

\citet{feng-etal-2021-survey} suggest that ``the distribution of augmented data should neither be too similar nor too different from the original''. FieldSwap achieves this balance by placing known key phrases in the contexts of other key phrases, which increases diversity in a controlled way. Data for underrepresented fields can be extrapolated using patterns learned from multi-shot fields \cite{lee2021neural}, but our work focuses on the scenario in which all fields are few-shot.
The use of schema field types in FieldSwap is similar to the use of entity types for mention replacement in named entity recognition, which is effective especially in low-data settings \cite{dai-adel-2020-analysis}.
Generating data for intermediate tasks (for example, natural language inference), especially when combined with self-training, is effective on a wide array of natural language processing tasks in the few-shot setting but assumes access to unlabeled data \cite{vu-etal-2021-strata}.

Other data augmentation techniques have been used for multimodal tasks that combine text and vision, such as image captioning \cite{app10175978} and visual question answering \cite{kafle-etal-2017-data,yokota-nakayama-2018-augmenting}. FieldSwap, like these other approaches, focuses on modifying the textual component of each input rather than the visual component; that is, the key phrase is replaced but the spatial layout remains the same.

Perhaps the most similar prior work to ours is \cite{andreas-2020-good}. The main idea of that work is that if two items appear in similar contexts, then they can be interchanged wherever one of them occurs to generate new examples. In our work, the items we change are key phrases associated with schema fields, and we determine interchangeability based on the identity or type of the field. Rather than generate new labeled examples by changing the value of the field, we generate examples by changing the surrounding context (via key phrases).

\subsection{Few-Shot Learning}
Because recent approaches to form extraction require substantial amounts of data, we are interested in improving performance in the few-shot setting. \citet{gunel2022data} demonstrate that transfer learning, especially multi-domain transfer learning, can greatly improve form extraction performance in the few-shot setting. A similar method for fine-tuning with data from other domains has been shown to work for other information extraction tasks \cite{xu-etal-2021-gradual}. The FieldSwap method we propose does not rely on existing data in other domains.

\subsection{Form Extraction}
Approaches for extracting information from form-like documents typically rely on multimodal features: text, spatial layout, and visual patterns. Models often make use of pre-trained encoders that incorporate such multimodal signals \cite{9710059,10.1007/978-3-030-86549-8_34,huang2022layoutlmv3}, but these encoders require a large amount of pre-training data, although they do exhibit good downstream task data efficiency during fine-tuning \cite{sage2021data}. Large amounts of training data are also required by span classification approaches \cite{majumder-etal-2020-representation,tata2021glean}, sequence labeling approaches \cite{aggarwal-etal-2020-form2seq,lee-etal-2022-formnet}, and end-to-end approaches \cite{cheng2022trie++}. Rather than suggest a new model architecture, we propose a method for augmenting form extraction data.
\section{Conclusions}
In this paper, we describe a data-augmentation technique designed for extraction problems on visually rich documents. We exploit the fact that many fields have a ``key phrase'' to indicate them: we generate an augmented example for a target field by taking an example of a source field and replacing the key phrase with that of the target field (while retaining the other neighboring tokens). 
Experiments on a variety of datasets show that this simple technique is very effective, particularly for small training sets (10--250 documents). 
Results show improvements of 1--7 points on the average macro F1 score, and up to 22 F1 points on some fields.

This result opens up two interesting directions for future work. First, how do we design a version of FieldSwap that works for sequence-labeling based modeling approaches given the challenges we described in Section~\ref{sec:field-swap}? Second, there are several extensions to FieldSwap that are worth investigating. When does swapping across document types help? Can we use a pre-trained LLM instead of a human expert to generate a set of key phrases given the name or description of a field? Can we learn information about key phrases from an unlabeled corpus to enable semi-supervised learning \cite{pryzant2022automatic}?

\balance


\bibliographystyle{ACM-Reference-Format}
\bibliography{references}

\appendix

\end{document}